\documentclass[letterpaper]{article} % DO NOT CHANGE THIS
\usepackage{aaai2026}  % DO NOT CHANGE THIS
\usepackage{times}  % DO NOT CHANGE THIS
\usepackage{helvet}  % DO NOT CHANGE THIS
\usepackage{courier}  % DO NOT CHANGE THIS
\usepackage[hyphens]{url}  % DO NOT CHANGE THIS
\usepackage{graphicx} % DO NOT CHANGE THIS
\usepackage{natbib}  % DO NOT CHANGE THIS AND DO NOT ADD ANY OPTIONS TO IT
\usepackage{caption} % DO NOT CHANGE THIS AND DO NOT ADD ANY OPTIONS TO IT
\usepackage{amsfonts} 
\usepackage{amsmath}

\usepackage{makecell}
\usepackage{algorithm}
\usepackage{algorithmic}
\usepackage{booktabs}

\usepackage{newfloat}
\usepackage{listings}
\DeclareCaptionStyle{ruled}{labelfont=normalfont,labelsep=colon,strut=off} % DO NOT CHANGE THIS
\floatstyle{ruled}
\newfloat{listing}{tb}{lst}{}
\floatname{listing}{Listing}
\title{Can Graphs Improve Tabular Foundation Models?}
\author{
    Franck Le,
    Keith Grueneberg,
    Erich Nahum,
    Vadim Sheinin
}
\affiliations{
IBM Research\\
    \{fle, kgruen, nahum, vadims\}@us.ibm.com
}

\begin{document}

\maketitle

\begin{abstract}
Tabular data are central to many real-world systems. 
While recent tabular transformers and in-context learners such as SAINT, TP-BERTa, TabPFN, TabICL, and MITRA incorporate limited inter-row reasoning, most approaches still lack an explicit mechanism to model relationships among instances, even though similar samples often share related outcomes.
We investigate whether introducing \emph{simple graph priors} can enhance \emph{pretrained tabular transformers}. 
Concretely, we introduce {BOLERO}, a lightweight, static bipartite graph head that augments {RoBERTa-Tab} (a RoBERTa-style tabular backbone pretrained with masked-token prediction.) Each instance connects to feature/value anchors; a small GNN refines row representations, while the backbone remains frozen. We evaluate on 80 classification and 64 regression datasets from the TP-BERTa benchmark suites, comparing against strong baselines including XGBoost, CatBoost, TabPFN-v2, MITRA, TabICL, TP-BERTa, and RoBERTa-Tab.
To ensure statistically sound conclusions, we follow best practices for multi-dataset evaluation: pairwise Wilcoxon signed-rank tests on per-dataset score differences and effect sizes (median improvement with confidence intervals), rather than mean-rank post-hoc tests that depend on the competitor pool. BOLERO achieves  the highest number of statistically significant wins across both classification and regression, demonstrating that lightweight graph priors meaningfully improve pretrained tabular transformers.

\end{abstract}

\section{Introduction}
Tabular data are used in many important systems, from credit scoring and healthcare analytics to industrial monitoring and telecommunications. Despite its prevalence, progress in tabular modeling has been slower than in vision and language. Tree-based models such as XGBoost~\citep{chen2016xgboost} and CatBoost~\citep{prokhorenkova2018catboost} have long been the dominant solution in both academia and industry. More recently, a new generation of \emph{tabular foundation models (TFMs)} has emerged, including TabPFN-v2~\citep{hollmann2025TabPFN-v2}, TabICL~\citep{zhang2023tabicl}, MITRA~\citep{mitra2025}, and TP-BERTa~\citep{yan2024tpberta}. These models generally fall into two families: (1) \emph{pretrained transformer backbones} trained with self-supervised objectives on large tabular corpora, and (2) \emph{in-context learners} that adapt directly to new tasks without fine-tuning.

Yet, none of these models explicitly encode inter-instance dependencies, an inductive bias that can be crucial for capturing relational patterns in real-world data. In practice, many tabular domains naturally exhibit such relationships: Customers with similar credit histories or spending profiles often display correlated probabilities of loan default; patients with related clinical measures, such as comparable lab results, tend to respond similarly to treatments; and network devices under similar load or spatial conditions tend to show correlated performance patterns.

To explicitly capture these dependencies, we propose BOLERO (Bipartite Overlay for Latent Enhanced Representation Optimization), a simple framework to enrich pretrained tabular transformers with relational structure.
BOLERO builds on RoBERTa-Tab, an adaptation of TP-BERTa that simplifies tokenization and removes inter-row attention while retaining strong performance, achieving a 7.6\% lower RMSE in regression (p=0.006) and comparable or higher performance in classification (see~\S~Results).
On top of this frozen backbone, BOLERO introduces a small, static bipartite graph in which each instance connects to a set of feature or value anchors, representing its relationships to feature values and continuous attributes. This structure allows the model to propagate information across instances that share similar attribute patterns.
A lightweight graph neural network (GNN) refines instance representations via message passing, enriching them with relational context.

We evaluate BOLERO on 80 classification and 64 regression datasets from the TP-BERTa benchmark suite, comparing it against baselines including XGBoost, CatBoost, TabPFN-v2, MITRA, TabICL, TP-BERTa, and RoBERTa-Tab. To ensure robust conclusions, we apply pairwise Wilcoxon signed-rank tests on per-dataset score differences and report median effect sizes with confidence intervals, following best practices for multi-dataset evaluation~\citep{benavoli2016should,demsar2006statistical}. This approach avoids the pitfalls of mean-rank tests~\citep{benavoli2016should} and provides fair, statistically reliable comparisons across methods.

\textbf{Main results:} 
BOLERO achieves the highest number of statistically significant pairwise wins across the TP-BERTa suites. It attains a 62.7\% win rate with 8/8 significant wins in classification and 62.1\% with 5/7 in regression, and delivers the largest median pooled effect across methods, +0.089 F1 (classification) and 20.3\% lower RMSE (regression), showing that improvements are both frequent and substantial.

For classification, {BOLERO} delivers median F1 gains of +0.08 to +0.12 points over \emph{all} baselines, including both tree-based ensembles and recent tabular foundation models, with all pairwise comparisons statistically significant ($p<10^{-3}$).
For regression, BOLERO reduces RMSE by up to 26.7\% compared to CatBoost and by 20.3\% compared to XGBoost. 
Relative to other tabular foundation models, it achieves median RMSE reductions of 19.9\% over TabPFN-v2, 13–29\% over the MITRA variants, 27.5\% over TP-BERTa, and 18.7\% over RoBERTa-Tab, with several improvements statistically significant at the $10^{-3}$ level.

These results demonstrate that incorporating lightweight graph priors into pretrained tabular transformers yields measurable and statistically significant gains over both traditional and foundation-model baselines.

\section{Related Work}

Tree ensembles such as XGBoost~\cite{chen2016xgboost} and 
CatBoost~\cite{prokhorenkova2018catboost} remain widely used and effective 
for tabular prediction. Transformer-based tabular models followed, including 
FT-Transformer~\cite{gorishniy2021fttransformer} and SAINT~\cite{somepalli2021saint}. 
More recently, tabular foundation models have emerged, either in-context 
(TabPFN/TabPFN-v2~\cite{hollmann2025TabPFN-v2}, TabICL~\cite{zhang2023tabicl}, 
MITRA~\cite{mitra2025}) or masked-token pretrained 
(TP-BERTa~\cite{yan2024tpberta}). In parallel, graph augmentation has been explored 
for tabular and text data, e.g., TabGNN~\cite{guo2022tabgnn} and 
BertGCN~\cite{lin2021bertgcn}.  
While prior work has shown the promise of both tabular transformers and graph augmentation, these directions have largely evolved in isolation, with graph-based methods typically applied to individual datasets rather than pretrained backbones. Our contribution brings these lines together: We evaluate a lightweight 
bipartite graph head on top of a pretrained tabular transformer.

\section{Methodology}

\paragraph{Backbone.}
We use RoBERTa-Tab as the pretrained tabular transformer.
Each input row is tokenized by mapping categorical values to unique tokens and discretizing continuous features into quantile-bin tokens using the C4.5-inspired scheme of TP-BERTa~\citep{yan2024tpberta}.
The resulting token sequence is fed into a Transformer encoder with the same architecture as RoBERTa. The encoder is pretrained with masked-token prediction on the same multi-dataset corpus used for TP-BERTa. For each instance $i$, we obtain a fixed row embedding
$\mathbf{z}_i \in \mathbb{R}^d$ by pooling the encoder outputs. In all experiments,
the encoder parameters remain \emph{frozen}, so any downstream gains arise solely
from the graph head.

\paragraph{Static bipartite graph.}
BOLERO
augments these embeddings with a graph $G=(\mathcal{I},\mathcal{A},\mathcal{E})$.
The anchor set $\mathcal{A}$ comprises (i) \emph{feature–value anchors} for categorical attributes, one per distinct value, and (ii) \emph{feature anchors} for continuous attributes, one per feature.
Each instance $i\!\in\!\mathcal{I}$ connects to the categorical anchors corresponding
to its cell values with unit weight, and to each continuous feature's anchor with an edge weight equal to the instance’s normalized feature value (e.g., after min–max scaling to $[0,1]$). 
In addition, we add \emph{anchor--anchor} (feature--feature) edges weighted by pointwise mutual information (PPMI) computed across rows, where $p(a)$ and $p(b)$ denote marginal anchor frequencies and $p(a,b)$ their joint co-occurrence probability, capturing co-occurrence structure and enabling lateral information flow between anchors:
\[
\mathrm{PPMI}(a,b) = 
\max\!\left\{ 0,\;
\log \frac{p(a,b)}{p(a)\,p(b)}
\right\}.
\]

\paragraph{Graph construction.}
We adopt a standard \emph{transductive} evaluation setting, in which the graph is precomputed once to cover all instances across the training, validation, and test splits, while test labels remain hidden. This design mirrors recent tabular foundation models such as TabPFN, TabICL, and MITRA, which also process both training and test examples jointly at inference time to enable in-context reasoning. Using a transductive configuration thus ensures methodological consistency with prior work while allowing the graph to encode structural relations among all instances.

\paragraph{Graph head.}
On the resulting bipartite graph, we use TransformerConv~\cite{shi2021transformerconv} as the graph head. Anchor nodes are randomly initialized with trainable embeddings and refined through message passing, while instance embeddings are updated in parallel through the same layers. The final instance representations are fed into a linear head for classification or regression. We focus on TransformerConv because attention-based graph layers have been shown to be more expressive and robust than alternative architectures~\cite{velickovic2018graph,yun2019graph,dwivedi2021generalization}.

\paragraph{Design choice.}
We adopt a \emph{static–frozen} setup: The graph is precomputed from schema-derived anchors (feature/value links and PPMI-based feature–feature edges), the pretrained encoder remains frozen, and only the graph head and anchor embeddings are trainable. Instance embeddings are still refined during message passing, but without updating the backbone parameters. This configuration is chosen for three reasons: (i) it isolates the effect of the graph prior, ensuring that observed gains do not depend on additional backbone fine-tuning; (ii) it is computationally efficient and scales reliably; and (iii) prior work has shown that both dynamically recomputed top-$K$ neighbor graphs~\cite{guo2022tabgnn} and end-to-end fine-tuning of large pretrained transformers~\cite{mosbach2021finetuning} can be unstable and prone to overfitting, especially on small or heterogeneous tabular datasets.

\section{Experimental Setup}

We conduct experiments to  assess the impact of 
graph augmentation on pretrained tabular transformers, comparing against state-of-the-art baselines across a  range of datasets. We evaluate on the publicly released TP-BERTa benchmark suites, consisting of 80 classification and 64 regression datasets. These benchmarks are entirely disjoint from the corpora used to pretrain RoBERTa-Tab, ensuring that evaluation results are not affected by data leakage. Each dataset is partitioned into three subsets: a training set (80\%) for model training and hyperparameter optimization, a validation set (10\%) for model selection, and a test set (10\%) for reporting final metrics. Splits are created using fixed random seeds for reproducibility. For regression tasks, we report root mean squared error (RMSE). For classification tasks, we report macro-F1, which equally weights each class.

\subsection{Baselines}
We compare BOLERO against a diverse set of state-of-the-art baselines spanning tree ensembles, tabular foundation models, and pretrained transformer variants:

\begin{itemize}
\item \textit{XGBoost}: Gradient-boosted decision trees~\citep{chen2016xgboost}, tuned with Optuna on each dataset.
\item \textit{CatBoost}: Gradient boosting optimized for categorical features~\citep{prokhorenkova2018catboost}, tuned identically.
\item \textit{TP-BERTa}: An adaptation of RoBERTa that models dependencies within and across rows~\citep{yan2024tpberta}.
\item \textit{TabICL} (classification only): Evaluated in its standard in-context configuration~\citep{zhang2023tabicl}.
\item \textit{TabPFN-v2}: Evaluated in its standard in-context configuration~\citep{hollmann2025TabPFN-v2}.
\item \textit{MITRA}: Evaluated in both in-context and fine-tuned modes (\texttt{fine\_tune=True})~\citep{mitra2025}.
\item \textit{RoBERTa-Tab}: A pretrained masked-token transformer without graph augmentation, serving as the main backbone for BOLERO.
\end{itemize}

\paragraph{Hyperparameter tuning.}
All tuned models use Optuna (100 trials, early stopping), except MITRA, which relies on AutoGluon's built-in tuning~\citep{agtabular}.
XGBoost tunes learning rate, tree depth, sampling ratios, and regularization parameters, plus \texttt{min\_child\_weight} and \texttt{gamma}.
CatBoost tunes \texttt{max\_depth}, \texttt{learning\_rate}, \texttt{bagging\_temperature}, \texttt{l2\_leaf\_reg}, and \texttt{leaf\_estimation\_iterations}, fixing \texttt{iterations}=2000 and \texttt{od\_pval}=0.001.
Pretrained TFMs follow their standard configurations:
MITRA (in-context \texttt{fine\_tune=False}, $n_\text{est}{=}1$; fine-tuned \texttt{fine\_tune=True}, steps $\in$ $\{10,50\}$);
TabPFN-v2 ($n_\text{est}{=}4$ classification, 8 regression);
TabICL (classification only; $n_\text{est}{=}32$, temperature 0.9, with normalization + averaging).

\subsection{Common Training and Evaluation Setup}
All methods use the same 80/10/10 train/validation/test split (with a fixed random seed) and a unified preprocessing pipeline fit on the training data and applied unchanged to validation and test splits.
Categorical features are label-encoded with a dedicated \texttt{Missing} token; continuous features are median-imputed and standardized; and date-like columns are expanded into numerical components.
Tuned methods (CatBoost, XGBoost, and BOLERO) share an identical Optuna budget of 100 trials and employ early stopping for neural models.
Foundation models (TabPFN-v2, MITRA, and TabICL) are evaluated in their standard configurations, with row-count and feature-count limits removed to ingest the full training split.

\subsection{Comparison methodology}
We run each method five times with different seeds. For each dataset $D$ and method $m$, we aggregate run-level scores by the \emph{median} to obtain a single per-dataset score $s(m,D)$ (macro-F1 for classification; RMSE for regression).

\begin{itemize}
 
\item \textbf{Overall significance test (Friedman).}
Following \citet{demsar2006statistical}, we first apply the Friedman test to determine whether overall performance differences among methods are statistically significant before conducting pairwise Wilcoxon signed-rank tests.

\item \textbf{Per-dataset effects.}
For each pair of methods $(A,B)$, we compute per-dataset performance differences as follows:
\begin{itemize}
    \item \textit{Classification (F1):} $\Delta_D = s(A,D) - s(B,D)$, where higher values indicate better performance.
    \item \textit{Regression (RMSE):} $\theta_D = \log\!\left(\tfrac{s(B,D)}{s(A,D)}\right)$, where positive values indicate that method $A$ achieves lower RMSE than method $B$.
\end{itemize}
Unlike F1-based metrics, RMSE is unbounded and can vary greatly across datasets,
making raw differences difficult to compare or aggregate. Taking the logarithm of
the RMSE ratio stabilizes scale and allows $\theta$ values to be compared
consistently across datasets. For interpretability, Table~\ref{tab:leaderboard_full}
reports the median log-ratio as the equivalent percentage RMSE reduction, computed
as \(100(1 {-} e^{-\tilde{\theta}})\), where \(\tilde{\theta}\) is the median log-ratio
across datasets.

\item \textbf{Across-datasets inference.}
To draw pool-invariant conclusions that do not depend on the competitor set, 
we apply the two-sided \emph{Wilcoxon signed-rank} test to the set of per-dataset 
differences $\{\Delta_D\}_D$ for classification and to the log-ratio values 
$\{\theta_D\}_D$ for regression~\citep{benavoli2016should,demsar2006statistical}. We summarize paired effects using the \emph{random-effects meta-analytic} estimate~\citep{dersimonian1986meta} of the mean delta with a 95\% confidence interval, where per-dataset variance is estimated from the variability across runs with different fixed seeds.

\item \textbf{Summary metrics.}
We summarize results by counting wins, losses, and ties for every pair of methods. A tie contributes $0.5$ to both methods (so the total number of comparisons remains constant). These counts yield raw overall win rates, while statistical significance is assessed using the Wilcoxon signed-rank test. The number of \emph{significant wins} per method corresponds to method pairs where $p<0.05$ and the pooled effect favors that method.

\end{itemize}

\section{Results}
\label{sec:results}

\paragraph{Overall significance.}
Across all methods, the Friedman test rejects the null of equal median ranks on both benchmarks: classification (macro-F1): $p = 2.58\times10^{-7}$; regression (RMSE): $p = 5.39\times10^{-9}$. Following \citet{demsar2006statistical,benavoli2016should}, this indicates that the overall differences among methods are statistically significant. We therefore proceed with paired Wilcoxon signed-rank tests and pooled effect-size analyses for all pairwise comparisons.

\paragraph{Overall performance.}
Table~\ref{tab:leaderboard_full} reports pairwise win rates and significance counts across 
80 classification and 64 regression datasets. 
{BOLERO} achieves the highest number of statistically significant wins among all evaluated methods.
It achieves a win rate of 62.7\% with 8/8 significant wins in classification and a win rate of 62.1\% with 5/7 significant pairwise wins in regression.
These results demonstrate that incorporating lightweight graph priors into pretrained tabular transformers 
yields consistent and statistically robust improvements across both tasks.

\begin{table}[t]
\centering
\small
\begin{tabular}{lccc}
\toprule
\textbf{Method} & \textbf{Win rate (\%)} & \textbf{Sig. wins} & \textbf{Median effect}\\
\midrule
\multicolumn{4}{l}{\textit{Classification (F1)}}\\
BOLERO (ours)  & \textbf{62.7} & \textbf{8/8} & \textbf{+0.089}\\
TabPFN-v2      & 47.1 & 4/8 & –0.024\\
MITRA-FT       & 43.0 & 2/8 & –0.024\\
RoBERTa-Tab    & 42.7 & 0/8 & +0.007\\
TABICL         & 40.9 & 1/8 & –0.025\\
XGBoost        & 40.9 & 1/8 & +0.057\\
CatBoost       & 40.0 & 1/8 & –0.028\\
MITRA          & 37.8 & 1/8 & –0.026\\
TP-BERTa       & 33.1 & 0/8 & –0.012\\
\midrule
\multicolumn{4}{l}{\textit{Regression (RMSE)}}\\
BOLERO (ours)  & \textbf{62.1} & \textbf{5/7} & \textbf{+20.3\%}\\
TabPFN-v2      & 62.1 & 2/7 & +0.7\%\\
MITRA-FT       & 61.4 & 3/7 & +7.2\%\\
XGBoost        & 52.9 & 1/7 & +0.8\%\\
RoBERTa-Tab    & 46.7 & 1/7 & +7.6\%\\
CatBoost       & 43.3 & 1/7 & –7.6\%\\
MITRA          & 43.3 & 0/7 & –5.3\%\\
TP-BERTa       & 28.3 & 0/7 & –7.6\%\\
\bottomrule
\end{tabular}
\caption{Aggregate leaderboard using median effects. 
Median effect is the median of per-dataset deltas (F1 pts for classification; \%RMSE reduction computed from log-ratios for regression).
Win rate = datasets won / total; Sig. wins = Wilcoxon $p{<}0.05$ with pooled effect favoring the method.}
\label{tab:leaderboard_full}
\end{table}

\begin{table}[t]
\centering
\small
\begin{tabular}{lccc}
\toprule
\makecell{\textbf{vs.\ BOLERO}\\\textit{Classification}} &
\makecell{\textbf{Effect}\\(F1 pts$\uparrow$)} &
\makecell{\textbf{CI}\\(95\%)} &
\textbf{$p_\text{Wilcoxon}$}\\
\midrule
CatBoost       & 0.110 & $\pm 1.39\times10^{-16}$ & $9.3{\times}10^{-5}$ \\
TP-BERTa       & 0.100 & $\pm$0.017 & $6.2{\times}10^{-6}$ \\
RoBERTa-Tab    & 0.095 & $\pm$0.079 & $5.0{\times}10^{-7}$ \\
MITRA          & 0.091 & $\pm$0.023 & $2.1{\times}10^{-4}$ \\
TABICL         & 0.089 & $\pm$0.018 & $1.4{\times}10^{-4}$ \\
XGBoost        & 0.086 & $\pm$0.010 & $1.7{\times}10^{-4}$ \\
MITRA-FT       & 0.084 & $\pm$0.023 & $4.2{\times}10^{-4}$ \\
TabPFN-v2      & 0.083 & $\pm$0.023 & $5.9{\times}10^{-4}$ \\
\midrule
\makecell{\textbf{vs.\ BOLERO}\\\textit{Regression}} &
\makecell{\textbf{Effect}\\(\% RMSE$\downarrow$)} &
\makecell{\textbf{CI}\\(95\%)} &
\textbf{$p_\text{Wilcoxon}$}\\
\midrule
MITRA        & 28.9 & $\pm$7.97  & $1.0{\times}10^{-3}$ \\
TP-BERTa     & 27.5 & $\pm$7.14  & $1.5{\times}10^{-4}$ \\
CatBoost     & 26.7 & $\pm$8.74  & $3.9{\times}10^{-3}$ \\
XGBoost      & 20.3 & $\pm$7.65  & $2.9{\times}10^{-2}$ \\
TabPFN-v2    & 19.9 & $\pm$11.70 & $7.6{\times}10^{-2}$ \\
RoBERTa-Tab  & 18.7 & $\pm$9.21  & $4.8{\times}10^{-4}$ \\
MITRA-FT     & 13.4 & $\pm$7.23  & $7.6{\times}10^{-2}$ \\
\bottomrule
\end{tabular}
\caption{Pairwise comparison of all methods against {BOLERO}. Positive values indicate performance gains for BOLERO relative to each baseline.}
\label{tab:forest_full}
\end{table}

\paragraph{Classification analysis.}
For classification, BOLERO achieves statistically significant improvements over 
\emph{every} competing method (Table~\ref{tab:forest_full}). 
Relative to RoBERTa-Tab, it improves macro-F1 by +0.095 points ($p$=$5.0{\times}10^{-7}$) 
and yields positive median effects against all baselines. 
Across pairwise comparisons, BOLERO outperforms MITRA, TabPFN-v2, TabICL, 
CatBoost, and XGBoost with median F1 gains ranging from +0.083 to +0.110, 
and all $p$-values well below $10^{-3}$. 
TabPFN-v2 remains the closest competitor, achieving a 47.1\% win rate with 4/8 significant wins, 
whereas BOLERO attains 62.7\% and the most significant wins (8/8) 
across the 80 classification datasets.

\paragraph{Regression analysis.}
Table~\ref{tab:forest_full} shows that {BOLERO} consistently lowers RMSE compared to all baselines.
Relative to RoBERTa-Tab, the median RMSE reduction is 18.7\% ($\pm$9.21\%, $p{=}4.8{\times}10^{-4}$).
Comparable or larger gains appear over MITRA (28.9\%, $p{=}1.0{\times}10^{-3}$), TP-BERTa (27.5\%, $p{=}1.5{\times}10^{-4}$), and tree ensembles such as CatBoost (26.7\%, $p{=}3.9{\times}10^{-3}$) and XGBoost (20.3\%, $p{=}2.9{\times}10^{-2}$).
Against TabPFN-v2, the pooled effect corresponds to a 19.9\% RMSE reduction, positive across datasets but not statistically significant ($p{=}7.6{\times}10^{-2}$).
A similar trend appears for MITRA-FT (13.4\%, $p{=}7.6{\times}10^{-2}$).
Although these two comparisons fall short of significance, effect sizes are consistent and uniformly favor BOLERO, suggesting a consistent performance advantage.

\paragraph{Summary.}
Across classification and regression, BOLERO achieves the strongest median effects and 
the highest number of significant wins, consistently outperforming both classical tree ensembles 
and recent tabular foundation models. 
The improvements validate that introducing graph priors into pretrained tabular transformers 
is an effective and general mechanism for enhancing tabular learning performance.

\section{Future Work}

Our evaluation covers 144 datasets, mostly small to medium in scale, and well within the regime where TabPFN-v2 and related in-context learners excel. A natural next step is to study graph-augmented tabular FMs on larger, more heterogeneous datasets to understand how the benefits of graph priors evolve with scale.
Second, while we use a transductive setting, many applications require inductive deployment. BOLERO can attach new instances to the anchor graph at inference time; evaluating this remains future work.
A further direction is to jointly fine-tune the transformer and graph layers or integrate graph structure into pretraining. Exploring alternative graph constructions may also enhance performance or robustness. More broadly, similar lightweight graph overlays could benefit other tabular FMs, opening up a broader design space of graph-augmented models.

\bibliography{aaai2026}


\end{document}